\documentclass{amsart}

\theoremstyle{definition}

\theoremstyle{remark}

\numberwithin{equation}{section}

\usepackage{amsmath,amssymb,amsfonts,amsrefs}
\usepackage{mathtools}
\usepackage{booktabs}
\usepackage{graphicx}
\usepackage[hidelinks]{hyperref}
\usepackage{xcolor}

\begin{document}

\title{Multiple Classifier for Concatenate-Designed Neural Network}

\author{Ka-Hou~Chan*}
\address{School of Applied Sciences, Macao Polytechnic Institute, Macao, China}
\email{chankahou@ipm.eud.mo}
\thanks{*Corresponding Author}

\author{Sio-Kei~Im}
\address{Macao Polytechnic Institute, Macao, China}

\author{Wei~Ke}
\address{School of Applied Sciences, Macao Polytechnic Institute, Macao, China}
\email{wke@ipm.eud.mo}


\date{\today}

\begin{abstract}
This article introduces a multiple classifier method to improve the performance of concatenate-designed neural networks,
such as ResNet and DenseNet,
with the purpose to alleviate the pressure on the final classifier.
We give the design of the classifiers, which collects the features produced between the network sets,
and present the constituent layers and the activation function for the classifiers, to calculate the classification score of each classifier.
We use the $\mathit{L2}$ normalization method to obtain the classifier score instead of the $\mathit{Softmax}$ normalization.
We also determine the conditions that can enhance convergence.
As a result, the proposed classifiers are able to improve the accuracy in the experimental cases significantly,
and show that the method not only has better performance than the original models, but also produces faster convergence.
Moreover, our classifiers are general and can be applied to all classification related concatenate-designed network models.
\end{abstract}

\maketitle

\section{Introduction}
\label{sec:Introduction}

Image classification is one of main topics of neural networks,
starting from the success of AlexNet~\cite{krizhevsky2012imagenet}.
The availability of object classification can lay the foundation for advanced neural systems
and is of great significance in the research of perceiving media data, such as
face recognition~\cite{lawrence1997face,sun2015deepid3,ranjan2017all},
medicine image analysis~\cite{shen2017deep,jiang2010medical,nawaz2018classification} and
autonomous vehicles~\cite{tian2018deeptest,kim2015deep}, \textit{etc.}
As we know that image classification by using neural networks has good performance in over 90.0\% of cases,
but there is the challenge of how to overcome the left cases with significant variations
of poor of illumination, image blurring and occlusions~\cite{gong2014re}.
Therefore, the current methods are hardly completely applicable to those cases where errors are strictly intolerable,
for instance the autonomous vehicles often miss the traffic light when driving fast in the midnight.

\begin{figure} [h]
	\centering
	\includegraphics{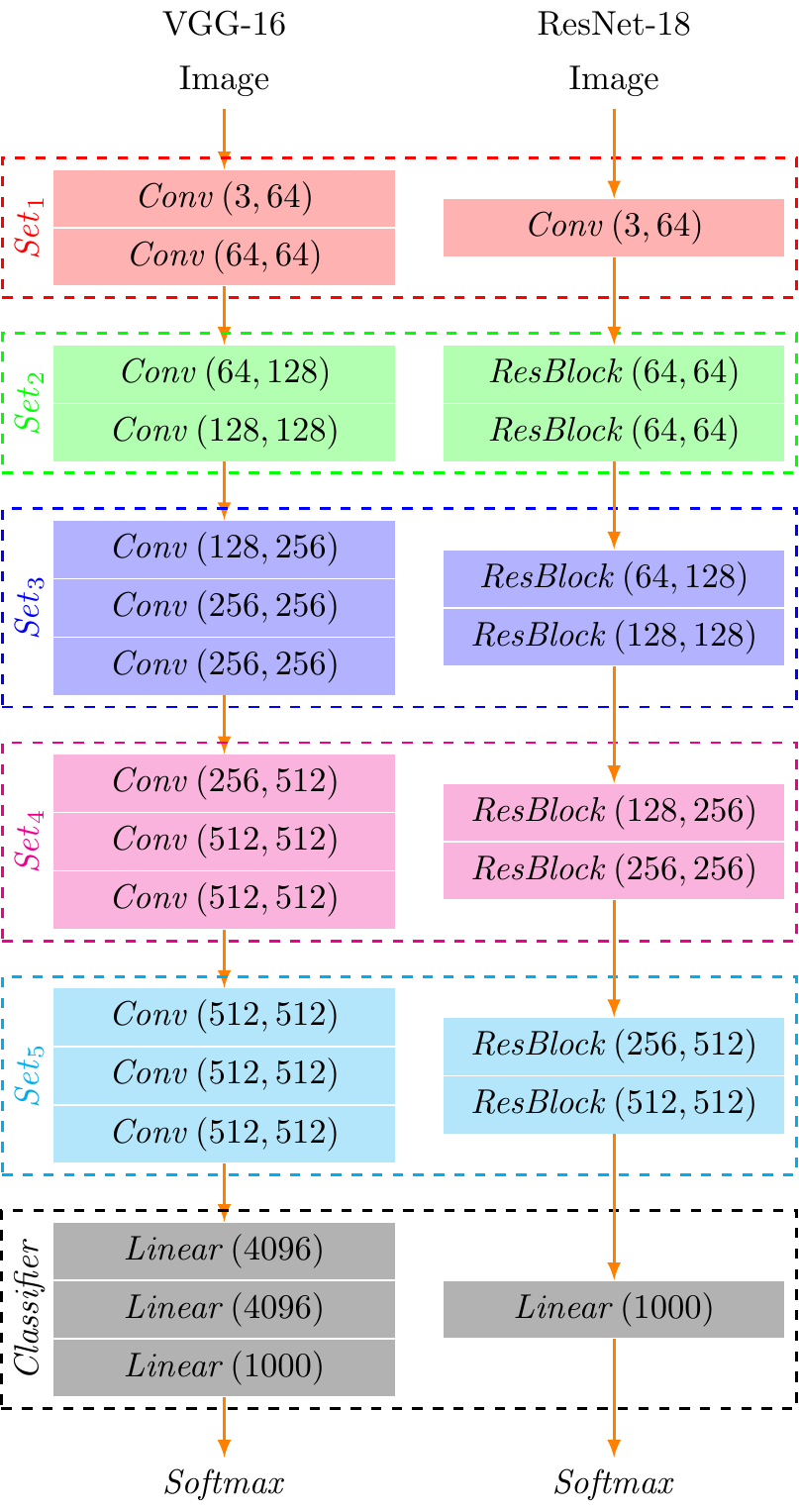}
	\caption{The brief architecture of VGG-16 and ResNet-18, both of the series has concatenated with five sets of stacked layers about $\mathit{Conv}$ and $\mathit{ResBlock}$, respectively, and one classifier is connected at the end of the model output.}
	\label{fig:BriefArchitecture}
\end{figure}

In order to increase the accuracy, studies on neural network architectures mostly focus on several aspects.
The easiest way is to improve the accuracy by stacking numerous layers,
but the increasing rate of this method presents a logarithmic curve that the later impact is very small.
Therefore, even if we increase the number of AlexNet's layers, the result is only slightly better than nothing~\cite{xiao2017scene,wang2015training}.
Another aspect considers images as the perceptual data that a learned model may describe random error or noise with redundancy,
instead of the strict underlying data distribution~\cite{zhang2016understanding}.
There are many data augmentation and regularization approaches proposed in the preprocessing,
such as random cropping~\cite{krizhevsky2012imagenet},
flipping~\cite{simonyan2014very} and
random erasing~\cite{zhong2020random}.
Data augmentation is closely related to oversampling in data analysis that can reduce overfitting when training machine learning models,
by applying data augmentation, one can achieve predictable accuracy improvement~\cite{shorten2019survey},
but it cannot overcome the shortcomings of the training model itself.
The most challenging aspect is to develop a new network model by combining layers strategically.
In order to obtain better performance, the state-of-the-art deeper network models, such as
GoogleNet~\cite{szegedy2015going},
ResNet~\cite{he2016deep} and
VGGNet~\cite{simonyan2014very}, \textit{etc.},
are designed with a large network architecture by stacking convolutional layers,
and these models have performed well for image classification.
With the in-depth study, it can be found that they all concatenate sets of similar convolutional layers together,
and connect one classifier at the end of the model output, see~\autoref{fig:BriefArchitecture}.
They are realized by feedforward neural networks, some features of similar images are always diluted during several convolutions.
For example, to determine the number `0' and the letter `O', their main features are similar that can be identified with the entire shape.
Considering the Convolutional Neural Networks~(CNNs), the convoluted results will only retain the local features
and discard the global information when pooling is performed with decreasing resolutions,
so the decision will be more difficult after the convolution and pooling.
However, no matter how good the fitting of the network design, the final classifier will be under great pressure to make decisions.

Below lists the main contribution of this paper.
\begin{itemize}
\item In order to alleviate the pressure on the final classifier,
we introduce multiple classifiers in this process, and then make decisions based on their results.

\item We present the constituent layers and activation function of the proposed classifiers,
the purpose is to calculate the classification scores of each classifier.

\item We introduce the $\mathit{L2}$ normalization method to obtain the classifier scores instead of using the $\mathit{Softmax}$ normalization,
and determine the conditions under which the convergence can be improved.

\item The proposed method is applicable to the state-of-the-art network models and achieves comparable results.
\end{itemize}
In particular, our approach is compatible with all discovered classification neural networks,
and can further be extended to deeper concatenate-designed models.
We conducted a comprehensive experiment on the CIFAR dataset~\cite{krizhevsky2009learning} to show the accuracy of VGGNet with and without our approach.
We show that the accuracy is obviously better with faster convergence.
Similar phenomena are also shown on the ResNet, which indicates that the optimization is effective,
and our approach is not just suitable to a particular network model.

The rest of this paper is organized as follows.
\autoref{sec:RelatedWork} goes through the related work.
\autoref{sec:MultipleClassifiers} present the design details and justification of the multiple classifier performance.
\autoref{sec:ExperimentsResultDiscussion} shows the experimental results.
Finally, \autoref{sec:Conclusion} concludes the paper.

\section{Related Work}
\label{sec:RelatedWork}

The multiple classifier approach has attracted a lot of research attention
and has been widely used in many perceived media such as image classification and speech recognition, \textit{etc.}
Researchers have studied multiple classifiers from different aspects,
including the classifier design and the concurrent rules.
Our work can be related to these three aspects, classifier design, type of classifier outputs and architectures.

For the traditional design in neural networks, by combining linear layers and activation functions,
a classifier can have multiple output units
and categorizes a sample according to the class
whose corresponding output gives the highest value among the multiple outputs~\cite{bridle1990probabilistic}.
Also, \cite{dietterich1994solving} proposed an error-correcting output code method to provide redundancy.
Later, \cite{zhou2012coding} proposed an idea of using an additional single-layer perceptron neural network to enhance the error-correcting capabilities.
In particular, a CNN always includes a number of convolutional and pooling layers
which are optionally accompanied by fully connected linear design.
\cite{ciresan2011convolutional} proved the robustness of their classifier by constructing seven CNNs.
It allows to consider the average error rate obtained as the best results.
Moreover, Support vector machines~(SVM), considered as one of the strongest and robustest algorithm in machine learning, was created by~\cite{vapnik1999overview},
and made use of~\cite{joachims1998making}.
It has become a well-known approach exploited in many domains~\cite{byun2003survey,naranjo2017two},
such as pattern recognition, classification and image processing,
and all of them obtained the best performance.
Later, \cite{elleuch2016new} modified the CNN structure by replacing the output layer of the fully connected with an SVM classifier.
In recent years, \cite{hershey2017cnn} involved multiple classifiers into their proposed model,
then averaged all the sets of the classifier outputs.
In addition, in order to better realize the function of the classifier in CNN,
two state-of-art classifiers Random Forest~(RF) and Gradient Boosting~(GB) have also been applied to deep learning~\cite{ball2017comprehensive}.
Meanwhile, for the types of classifier outputs, researchers considered how to calculate the confidence of the classifier for each image category,
where each sample was represented by multiple images~\cite{mandelbaum2017distance}.
Each classifier must be trained to increase the confidence of the corresponding category,
thus the category can be determined using the confidence of each classifier~\cite{ueda2000optimal}.
Further, the results of the multiple classifiers can also be used to determine in late fusion in multi-modal~\cite{lai2015learning} or multi-voting~\cite{cao2018dynamic} classification.
Specifically, the above method has a set of individual classifiers,
each classifier makes a decision on the input set individually,
then the method combines their decisions to form a composite result~\cite{diez2015diversity}.
In order to invoke the classifier in a concatenate-designed model,
\cite{wu2019wider} proposed convolutions as classifiers, instead of linear classifiers at the end of ResNet.
Further, \cite{hammann2020identity} decided to combine the ``long-term dependencies'' and the ResNet networks in one classifier
to show that the accuracy was improved significantly while maintaining a suitable interference time.

Meanwhile, the deeper convolutional architecture was the most important work,
demonstrating the powerful functions of concatenate-designed neural networks,
which showed that building a deeper network with a tiny convolution kernel was effective in increasing the performance of the CNN-based network models.
After VGGNet~\cite{simonyan2014very}, ResNet was first proposed by~\cite{he2016deep}.
It greatly alleviated the optimization difficulty and increased it to some hundreds of layers
by using skipping connections within their convolutional set.
Since then, different kinds of inner structures have been proposed, concentrating on various tasks
and consistently achieved the better performance in different areas~\cite{yunpeng2017sharing,xie2017aggregated}.
Further, \cite{huang2017densely} introduced the DenseNet which passed the input features to the output through a densely connected path
to concatenate the input features with the convoluted output as the \textit{DenseBlock} result.
Take account to these networks design, they aim to retain more original information to the classifier,
so they also tend to connect the input features directly to the output.
However, the width of their connection path increases linearly as the depth rises,
causing the number of training parameters to increase seriously.
This limits the building of deeper and wider networks that might further improve the accuracy.

In this work, inspired by the connection from the original information to the classifier,
we aim to adopt the proposed classifier following each layer set.
These classifiers make interim decisions instead of one decision of the last classifier.
Based on these decisions, we then propose a novel combination method and add it to the state-of-the-art concatenate-designed network that can achieve higher accuracy.

\section{Multiple Classifier Strategy}
\label{sec:MultipleClassifiers}

Looking into concatenate-designed networks, there are several sets connected in order, as shown in \autoref{fig:BriefArchitecture},
composing multiple convolutional layers and various types of blocks.
We use $h_t$ to denote the output feature of $\mathit{Set}_t$ at the $t$-th step
with attributes $\left[\mathit{batch},\mathit{channel},\mathit{width},\mathit{height}\right]$,
and $h_0$ is the original image.
For each step, $\mathit{Set}_t$ refers to the feature extraction function
that takes the previous feature as input and output the extracted information,
\begin{equation} \label{eq:output_feature}
h_t = \mathit{Set}_t\left(h_{t-1}\right).
\end{equation}
Then, the classification function $\mathit{Classifier}$ transforms the last output feature $h_{-1}$ to a specific dimension vector $\vec{c}$,
\begin{equation} \label{eq:dimension_vector}
\vec{c} = \mathit{Classifier}\left(h_{-1}\right).
\end{equation}
We further transform $\vec{c}$ into a probability vector through the $\mathit{Softmax}$ function.
In \autoref{eq:output_feature} and \autoref{eq:dimension_vector},
we encapsulate the network rule of various concatenate-designed architectures in a generalized way.
This observation shows that the connection path is essentially an extensible higher order function that extracts information from the previous states.
However, the feature size is reduced and the number of channels is increased when a feature passes through the sets.
Although it can capture the major part of the classification target, the overall structure of the image is somewhat lost.
In addition, some useful information in the later sets may also be discarded during the extraction from the earlier sets.
These problems become more obvious as the number of connections increases.
Therefore, the concatenation usually connects up to five sets, and the final classifier will be under great pressure to make decisions.

In order to address this issue, let's revisit the network and divide it by the pooling layers.
In this view, the architecture can be considered as a simple CNN network if $t=1$,
similar to LeNet-5~\cite{lecun1998gradient} if $t=2$,
and similar to AlexNet~\cite{krizhevsky2012imagenet} if $t=3$, \textit{etc.}
Particularly, the most advanced CNN network also satisfies with the $t=5$ case~\cite{he2016deep,simonyan2014very,huang2017densely}.
Thus, with the advancement of hardware performance, CNN models in different eras can be summarized as this series of neural networks with different $t$ values.
From the above analysis, we observe that the number of sets can increase indefinitely.
In practical applications, we only consider the part before the classifier, \textit{i.e.},
$\exists t \bullet \mathit{Classifier}_t\left(h_t\right) \equiv \vec{c}$.
Based on this idea, we propose to employ the number of $t$ classifiers as
\begin{equation} \label{eq:proposed_c}
\forall t \bullet \sum\mathit{Classifier}_t\left(h_t\right)=\sum\vec{c}_t\equiv\vec{c}.
\end{equation}
Meanwhile, such a decision contributing strategy makes it possible to compromise the global structure and the local pattern of an image.
Comparatively, the former one gets more original information and the latter one obtains the extracted information.
All classifiers make decisions based on their recent acquired features independently.
This strategy can provide more references for decision making,
especially in controversial cases such as `0' and `O', `1' and `l'.
Multiple classifiers can alleviate the perplexity, leading to high redundancy.

\begin{figure} [h]
	\centering
	\includegraphics{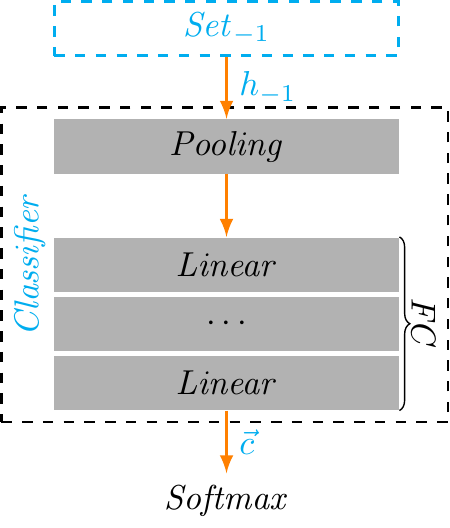}
	\caption{A general procedure of classifier: often consists of one $\mathit{Pooling}$ layer and multiple $\mathit{Linear}$ layers.
		 These {\em Linear} layers are often collectively referred to as Fully-Connected~(FC) layers.}
	\label{fig:classifierArch}
\end{figure}

\subsection{Classifier Design}
Before we present our classifier design,
we first analyze those classifiers used in the current neural networks.
It can be found that the procedure of their stacked layers is the same (see \autoref{fig:classifierArch}).
The classifier always receives the features produced by the last set, and must consider following dynamic attributes:
\begin{description}
\item[$\mathit{width} \times \mathit{height}$,] the size of the image varies with different applications,
but the Fully-Connected~(FC) layer only accepts a fixed size.
Classifiers have to pool a fix size in order to facilitate feature extraction from all sizes of $h_{-1}$.

\item[$channel$,] different from the image size, this attribute of the original image $h_0$ always equals to 3, corresponding to the RGB channels,
and then increases to 64, 128, 256 and 512 (up to 1024 in DenseNet) through the various predefined set.
\end{description}
Therefore, the number of input features of an FC layer is $\mathit{channel} \times \mathit{width} \times \mathit{height}$,
and a vector $\vec{c}$ is produced as the result.
According to this design, when applied to our classifier that satisfies \autoref{eq:proposed_c},
there are multiple classifiers to collect every output feature from $h_0$ to $h_{-1}$.
The number of classifiers is the same as the number of concatenated sets,
and we use $\mathit{Classifier}_t$ to denote the $t$-th classifier we employed.
Corresponding to the order of $\mathit{Set}_t$, the first problem is to fix the channel differences,
because the concatenation operation used in \autoref{eq:proposed_c} is not viable when the size of the features changes.
Thus, our classifier makes use of one convolutional layer to adjust the number of channel.
Regardless of the output feature $h_t$, we also aim to increase their channels to match the last feature $h_{-1}$
with a small convolutional kernel size $\left(3\times3\right)$,
each side of the input is zero-padded by one pixel to keep the feature size fixed.
Then we use an adaptive maximum pooling layer to downsample the convoluted results into a fixed size $\left(1\times1\right)$,
followed by a batch normalization~\cite{ioffe2015batch} as a transition layer between the pooling and FC layers.
Since the feature size has been downsampled to $\left(1\times1\right)$,
the number of FC inputs will equal to the size of the channel merely.
For simplicity, we only use one linear layer to achieve the FC transformation in this work,
followed by the type of rectifier activation function to ensure that the score is positive.
We do not recommend using the $\mathit{ReLU}$ because they become inactive for essentially all inputs smaller than zero.
In this state, no gradients flow backward through the neuron, and so the neuron falls into a permanent inactive state,
becoming a dead neuron, and will not be conducive to score any category.
In view of this, our design tends to use the $\mathit{Softplus}$ that can be viewed as a smooth version of $\mathit{ReLU}$,
which is monotonic and differentiable, having a positive first order derivative in $\mathbb{R}$.
By stacking the above layers, we have completed the feature extraction within the proposed classifiers,
and each classifier generates a score vector for each category.

\subsection{Score Normalization}
\label{subsec:ScoreNormalization}

In addition to feature extraction, we must also normalize the score vector to collect the result of each classifier.
We design that the classifier should provide the confidence rate (score) rather than making a classification decision.
The result is regarded as the scores for each category.
Usually, this part can be achieved directly through the $\mathit{Softmax}$ function, but the convergence is slow in practice,
so we introduce the method of using the $\mathit{L2}$ normalization to enhance the convergence of this part.

We first review the general form of the $\mathit{L1}$ and $\mathit{L2}$ normalization,
for any $f_1\left(x_i\right)$ and $f_2\left(x_i\right)$,
\begin{equation*}
\mathit{L1}\left(f_1\left(x_i\right)\right)=\frac{f_1\left(x_i\right)}{\sum_k f_1\left(x_k\right)},
\end{equation*}
\begin{equation*}
\mathit{L2}\left(f_2\left(x_i\right)\right)=\frac{f_2\left(x_i\right)}{\sqrt{\sum_k f^2_2\left(x_k\right)}},
\end{equation*}
where $N$ denote the number of categories, and the respect partial derivatives are,
\begin{equation*}
\frac{\partial\mathit{L1}\left(f_1\left(x_i\right)\right)}{\partial x_j}
=
\frac{1}{\sum_k f_1\left(x_k\right)}\left( - \frac{f_1\left(x_i\right)}{\sum_k f_1\left(x_k\right)}\right)\frac{\partial f_1\left(x_j\right)}{\partial x_j},
\end{equation*}
\begin{equation*}
\frac{\partial\mathit{L2}\left(f_2\left(x_i\right)\right)}{\partial x_j}
=
\frac{1}{\sqrt{\left(\sum_k f^2_2\left(x_k\right)\right)}}\left( - \frac{f_2\left(x_i\right)f_2\left(x_j\right)}{\sum_k f^2_2\left(x_k\right)}\right)\frac{\partial f_2\left(x_j\right)}{\partial x_j}.
\end{equation*}
It is worth noting that the $\mathit{L1}\left(f_1 \left(x\right)\right)$ becomes the $\mathit{Softmax}$ normalization if $f_1\left(x\right)=e^x$.
The normalization method we propose here is $\mathit{L2}\left(f_2\left(x\right)\right)$ with $f_2\left(x\right)=\sqrt{e^x}$.
We use $S\left(x\right)$ and $L\left(x\right)$ to denote $\mathit{Softmax}$ and the proposed normalization method, respectively.
According to above discussion, we formulate the final form as,
\begin{equation*}
S\left(x_i\right)=\mathit{L1}\left(e^{x_i}\right)=\frac{e^{x_i}}{\sum_k e^{x_k}},
\end{equation*}
\begin{equation*}
L\left(x_i\right)=\mathit{L2}\left(\sqrt{e^{x_i}}\right)=\sqrt{\frac{e^{x_i}}{\sum_k e^{x_k}}}.
\end{equation*}
The corresponding partial derivatives become,
\begin{equation} \label{eq:parS}
\frac{\partial S\left(x_i\right)}{\partial x_j}
=
\frac{1}{\sum_k e^{x_k}}\left(- \frac{e^{x_i}}{\sum_k e^{x_k}}\right)e^{x_j},
\end{equation}
\begin{equation} \label{eq:parL}
\frac{\partial L\left(x_i\right)}{\partial x_j}
=
\frac{1}{\sqrt{\sum_k e^{x_k}}}\left(- \frac{\sqrt{e^{x_i}} \sqrt{e^{x_j}}}{\sum_k e^{x_k}}\right)\frac{1}{2}\sqrt{e^{x_j}}.
\end{equation}
Since our goal is to enhance the convergence,
we must find a condition that satisfies the proposition about the partial derivatives,
$\frac{\partial L\left(x_i\right)}{\partial x_j} \geq \frac{\partial S\left(x_i\right)}{\partial x_j}$, \textit{i.e.},
\begin{equation*}
\frac{1}{\sqrt{\sum_k e^{x_k}}}\left(- \frac{\sqrt{e^{x_i}} \sqrt{e^{x_j}}}{\sum_k e^{x_k}}\right)\frac{1}{2}\sqrt{e^{x_j}}
\geq
\frac{1}{\sum_k e^{x_k}}\left(- \frac{e^{x_i}}{\sum_k e^{x_k}}\right)e^{x_j}.
\end{equation*}
We simplify this and obtain,
\begin{equation}\label{eq:dercon}
\sum_k e^{x_k} \leq 4e^{x_i}.
\end{equation}
It can be found that \autoref{eq:dercon} leads to the necessary condition $N<4e^{x_i}$, which can be rewritten as,
\begin{equation*}
x_i > \ln\frac{N}{4},
\end{equation*}
when we assume all $x_i>0$, because there is a $\mathit{Softplus}$ function at the end of the feature extraction. In summary, by using our normalization method with necessary condition $x_1,x_2,\cdots,x_k>\ln\frac{N}{4}$, the enhancement of convergence becomes more and more obvious with the increase of accuracy. Based on the theory, the proposed classifiers must have the lower bound $\ln\frac{N}{4}$. In fact, the $\mathit{Softplus}$ function always satisfies with $x>0\geq\ln\frac{N}{4}$ for $N\geq4$.

\begin{figure} [h]
\centering
\includegraphics{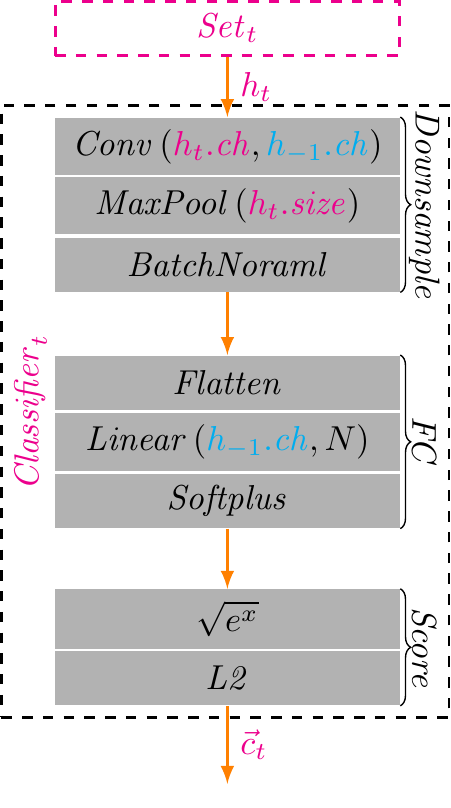}
\caption{The complete internal design of the proposed classifier.
    $h_t$ is the $t$-th output feature produced from $\mathit{Set}_t$,
    $h_{-1}$ is the last output feature, which is the same as in \autoref{fig:classifierArch},
    $h.\mathit{ch}$ and $h.\mathit{size}$ denote the number of $\mathit{channels}$ and the $\mathit{size}$ of feature $h$, respectively,
    $N$ denotes the number of categories.}
\label{fig:classififer}
\end{figure}

As shown in \autoref{fig:classififer}, we detail the structure of all the stacked layers in the classifier.
There are three parts of the procedures for each classifier.
First, $\mathit{Downsample}$ performs the feature extraction.
Next, $\mathit{FC}$ projects the extracted information into each category as a reference.
Finally, $\mathit{Score}$ normalizes these references into the output vector $\vec{c}_t$ for the final decision making with other classifiers.

\subsection{Network Architectures}

Following the last stage of the classification neural network,
we must finally provide a probability vector to calculate the cross entropy loss through the $\mathit{Softmax}$ function.
Considering the variable length of future concatenate-designed network,
the number of vectors $\vec{c}_t$ produced by our classifier that are used in other networks may also be different.
Therefore, we want to maintain its scalability and use the sum of the vectors in \autoref{eq:proposed_c}
as the final output of the multiple classifier strategy.
The overall design of the proposed method can inherit the backbone architecture of any concatenate-designed neural network,
making it easy to implement and apply to other tasks.
This can be achieved simply by adding a classifier following each $\mathit{Set}_t$ to the existing classification networks.
Under a well optimized deep learning platform, each classifier requires only a fixed amount of computational cost and memory consumption,
making the deployment very efficient.

\begin{figure} [h]
\centering
\includegraphics{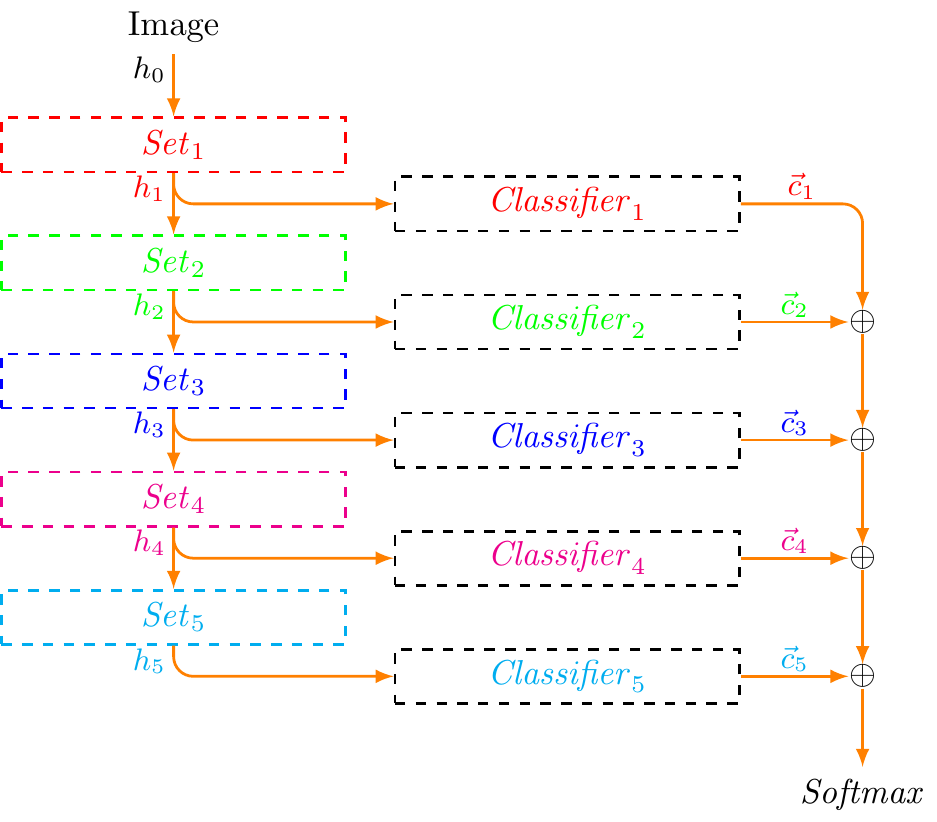}
\caption{The complete structure of proposed classification method.
    $\mathit{Set}_t$ is contributed by the original network design (like \autoref{fig:BriefArchitecture}).
    The proposed $\mathit{Classifier}_t$ collects output feature $h_t$,
    and then produces a vector $\vec{c}_t$ as a reference for final decision, instead of using a classifier at the end.}
\label{fig:NN}
\end{figure}

\begin{table} [h]
\centering
\caption{The architecture and complexity of our re-implemented concatenate-designed neural networks.
    We present their required training parameters and computational cost using the FLOPS with input size of one $3\times32\times32$ image.}
\begin{tabular}{c|c|c|c|c}
\toprule
\textbf{\textit{Stage}} & \textbf{VGG16} & \textbf{ResNet18} & \textbf{DLA34} & \textbf{DenseNet121} \\
\midrule
$\mathit{Set}_1$ & $\begin{bmatrix}3\times3,64\end{bmatrix}\times2$ & $\begin{bmatrix}3\times3,64\end{bmatrix}\times1$ & $\begin{bmatrix}3\times3,16\end{bmatrix}\times1$ & $\begin{bmatrix}3\times3,64\end{bmatrix}\times1$ \\
\midrule
$\mathit{Set}_2$ & $\begin{bmatrix}3\times3,128\end{bmatrix}\times2$ & $\begin{bmatrix}3\times3,64\\3\times3,64\end{bmatrix}\times2$ & $\begin{bmatrix}3\times3,32\end{bmatrix}\times1$ &
$\begin{matrix}\begin{bmatrix}1\times1,128\\3\times3,32\end{bmatrix}\times6\\\midrule\begin{bmatrix}1\times1,128\end{bmatrix}\times1\end{matrix}$ \\
\midrule
$\mathit{Set}_3$ & $\begin{bmatrix}3\times3,256\end{bmatrix}\times3$ & $\begin{bmatrix}3\times3,128\\3\times3,128\end{bmatrix}\times2$ & $\begin{matrix}\begin{bmatrix}3\times3,64\\3\times3,64\end{bmatrix}\times2\\\midrule\begin{bmatrix}1\times1,64\end{bmatrix}\times1\end{matrix}$ & $\begin{matrix}\begin{bmatrix}1\times1,128\\3\times3,32\end{bmatrix}\times12\\\midrule\begin{bmatrix}1\times1,256\end{bmatrix}\times1\end{matrix}$ \\
\midrule
$\mathit{Set}_4$ & $\begin{bmatrix}3\times3,512\end{bmatrix}\times3$ & $\begin{bmatrix}3\times3,256\\3\times3,256\end{bmatrix}\times2$ & $\begin{matrix}\begin{array}{c|c}\begin{matrix}\begin{bmatrix}3\times3,128\\3\times3,128\end{bmatrix}\times2\\\midrule\begin{bmatrix}1\times1,128\end{bmatrix}\times1\end{matrix}&\times2\end{array}\\\midrule\begin{bmatrix}1\times1,128\end{bmatrix}\times1\end{matrix}$ & $\begin{matrix}\begin{bmatrix}1\times1,128\\3\times3,32\end{bmatrix}\times24\\\midrule\begin{bmatrix}1\times1,512\end{bmatrix}\times1\end{matrix}$ \\
\midrule
$\mathit{Set}_5$ & $\begin{bmatrix}3\times3,512\end{bmatrix}\times3$ & $\begin{bmatrix}3\times3,512\\3\times3,512\end{bmatrix}\times2$ & $\begin{matrix}\begin{array}{c|c}\begin{matrix}\begin{bmatrix}3\times3,256\\3\times3,256\end{bmatrix}\times2\\\midrule\begin{bmatrix}1\times1,256\end{bmatrix}\times1\end{matrix}&\times2\end{array}\\\midrule\begin{bmatrix}1\times1,256\end{bmatrix}\times1\end{matrix}$& $\begin{bmatrix}1\times1,128\\3\times3,32\end{bmatrix}\times16$ \\
\midrule
$\mathit{Set}_6$ & & & $\begin{matrix}\begin{bmatrix}3\times3,512\\3\times3,512\end{bmatrix}\times2\\\midrule\begin{bmatrix}1\times1,512\end{bmatrix}\times1\end{matrix}$ \\
\midrule
$\mathit{Classifier}$ & $\begin{bmatrix}512\times4096\\4096\times4096\\4096\times N\end{bmatrix}$ & $\begin{bmatrix}512\times N\end{bmatrix}$ & $\begin{bmatrix}512\times N\end{bmatrix}$ & $\begin{bmatrix}1024\times N\end{bmatrix}$ \\
\midrule
\midrule
Training parameters & $15.7\times10^6$ & $10.2\times10^6$ & $15.1\times10^6$ & $18.0\times10^6$ \\
\midrule
FLOPS & $0.3\times10^9$ & $0.9\times10^9$ & $0.3\times10^9$ & $1.1\times10^9$\\
\bottomrule
\end{tabular}
\label{tab:original}
\end{table}

\begin{table} [h]
\centering
\caption{The architecture and complexity of our re-implemented concatenate-designed neural networks with the proposed multiple classifier strategy.
    We present their required training parameters and computational cost using the FLOPs with input size of one $3\times32\times32$ image.}
\begin{tabular}{c|c|c|c|c}
\toprule
\textbf{\textit{Stage}} & \textbf{VGG16} & \textbf{ResNet18} & \textbf{DLA34} & \textbf{DenseNet121} \\
\midrule
$\mathit{Set}_1$ & $\begin{bmatrix}3\times3,64\end{bmatrix}\times2$ & $\begin{bmatrix}3\times3,64\end{bmatrix}\times1$ & $\begin{bmatrix}3\times3,16\end{bmatrix}\times1$ & $\begin{bmatrix}3\times3,64\end{bmatrix}\times1$ \\
\midrule
$\mathit{Set}_2$ & $\begin{bmatrix}3\times3,128\end{bmatrix}\times2$ & $\begin{bmatrix}3\times3,64\\3\times3,64\end{bmatrix}\times2$ & $\begin{bmatrix}3\times3,32\end{bmatrix}\times1$ &
$\begin{matrix}\begin{bmatrix}1\times1,128\\3\times3,32\end{bmatrix}\times6\\\midrule\begin{bmatrix}1\times1,128\end{bmatrix}\times1\end{matrix}$ \\
\midrule
$\mathit{Set}_3$ & $\begin{bmatrix}3\times3,256\end{bmatrix}\times3$ & $\begin{bmatrix}3\times3,128\\3\times3,128\end{bmatrix}\times2$ & $\begin{matrix}\begin{bmatrix}3\times3,64\\3\times3,64\end{bmatrix}\times2\\\midrule\begin{bmatrix}1\times1,64\end{bmatrix}\times1\end{matrix}$ & $\begin{matrix}\begin{bmatrix}1\times1,128\\3\times3,32\end{bmatrix}\times12\\\midrule\begin{bmatrix}1\times1,256\end{bmatrix}\times1\end{matrix}$ \\
\midrule
$\mathit{Set}_4$ & $\begin{bmatrix}3\times3,512\end{bmatrix}\times3$ & $\begin{bmatrix}3\times3,256\\3\times3,256\end{bmatrix}\times2$ & $\begin{matrix}\begin{array}{c|c}\begin{matrix}\begin{bmatrix}3\times3,128\\3\times3,128\end{bmatrix}\times2\\\midrule\begin{bmatrix}1\times1,128\end{bmatrix}\times1\end{matrix}&\times2\end{array}\\\midrule\begin{bmatrix}1\times1,128\end{bmatrix}\times1\end{matrix}$ & $\begin{matrix}\begin{bmatrix}1\times1,128\\3\times3,32\end{bmatrix}\times24\\\midrule\begin{bmatrix}1\times1,512\end{bmatrix}\times1\end{matrix}$ \\
\midrule
$\mathit{Set}_5$ & $\begin{bmatrix}3\times3,512\end{bmatrix}\times3$ & $\begin{bmatrix}3\times3,512\\3\times3,512\end{bmatrix}\times2$ & $\begin{matrix}\begin{array}{c|c}\begin{matrix}\begin{bmatrix}3\times3,256\\3\times3,256\end{bmatrix}\times2\\\midrule\begin{bmatrix}1\times1,256\end{bmatrix}\times1\end{matrix}&\times2\end{array}\\\midrule\begin{bmatrix}1\times1,256\end{bmatrix}\times1\end{matrix}$& $\begin{bmatrix}1\times1,128\\3\times3,32\end{bmatrix}\times16$ \\
\midrule
$\mathit{Set}_6$ & & & $\begin{matrix}\begin{bmatrix}3\times3,512\\3\times3,512\end{bmatrix}\times2\\\midrule\begin{bmatrix}1\times1,512\end{bmatrix}\times1\end{matrix}$ \\
\midrule
$\mathit{Proposed\,Classifier}$ & $\begin{bmatrix}3\times3,512\\512\times N\end{bmatrix}\times5$ & $\begin{bmatrix}3\times3,512\\512\times N\end{bmatrix}\times5$ & $\begin{bmatrix}3\times3,512\\512\times N\end{bmatrix}\times6$ & $\begin{bmatrix}3\times3,1024\\1024\times N\end{bmatrix}\times5$ \\
\midrule
\midrule
Training parameters & $21.5\times10^6$ & $15.9\times10^6$ & $20.0\times10^6$ & $26.3\times10^6$ \\
\midrule
FLOPS & $0.5\times10^9$ & $1.4\times10^9$ & $0.6\times10^9$ & $2.0\times10^9$\\
\bottomrule
\end{tabular}
\label{tab:MC}
\end{table}

As listed in \autoref{tab:original} and \autoref{tab:MC},
we measure the model complexity by counting the total number of training parameters within each neural network,
and measure the computational cost of each deeper model using the floating-point operations per second~(FLOPS).
As found in the results, the required parameters of the neural network when using the proposed multiple classifiers
are about 50.0\% more than that of the original network, and the FLOPS reaches up to 200.0\% in a training epoch.

\section{Experiments Result and Discussion}
\label{sec:ExperimentsResultDiscussion}

In order to evaluate the proposed method on a variety of state-of-art classification models,
we applied our approach to the CIFAR-10 and CIFAR-100 datasets~\cite{krizhevsky2009learning},
which are labeled by 10 and 100 classes color images, respectively, with 50k for training and 10k for testing.
All our experiments are conducted on an NVIDIA GeForce RTX 2080 Ti with 11.0GB of video memory.
In order to compare the proposed strategy with the original networks,
we re-implement the VGG16~\cite{simonyan2014very}, ResNet18~\cite{he2016deep}, DLA34~\cite{yu2018deep} and DenseNet121~\cite{huang2017densely}
with and without our multiple classifiers in PyTorch~\cite{paszke2017automatic}, respectively,
and used the advanced gradient-related optimizer, the Adam~\cite{kingma2014adam} method with a learning rate of 0.001.
All experiments use the same dataset in each test with a batch size of 100 per iteration set,
and with the same configuration and the same number of neural nodes, as shown in \autoref{tab:original} and \autoref{tab:MC}.
For a justified comparison, we also train the original and the proposed models with the same training procedure.
We use random cropping and horizontal flipping with color normalization,
followed by the random erasure data augmentation~\cite{zhong2020random}.
We target to process 300 epochs to compare the accuracy and convergence of the various models.
In addition, there is a scheduler for adjusting the learning rate.
It reduces the learning rate when the loss becomes stagnant.

\begin{figure} [h]
	\centering
	\includegraphics{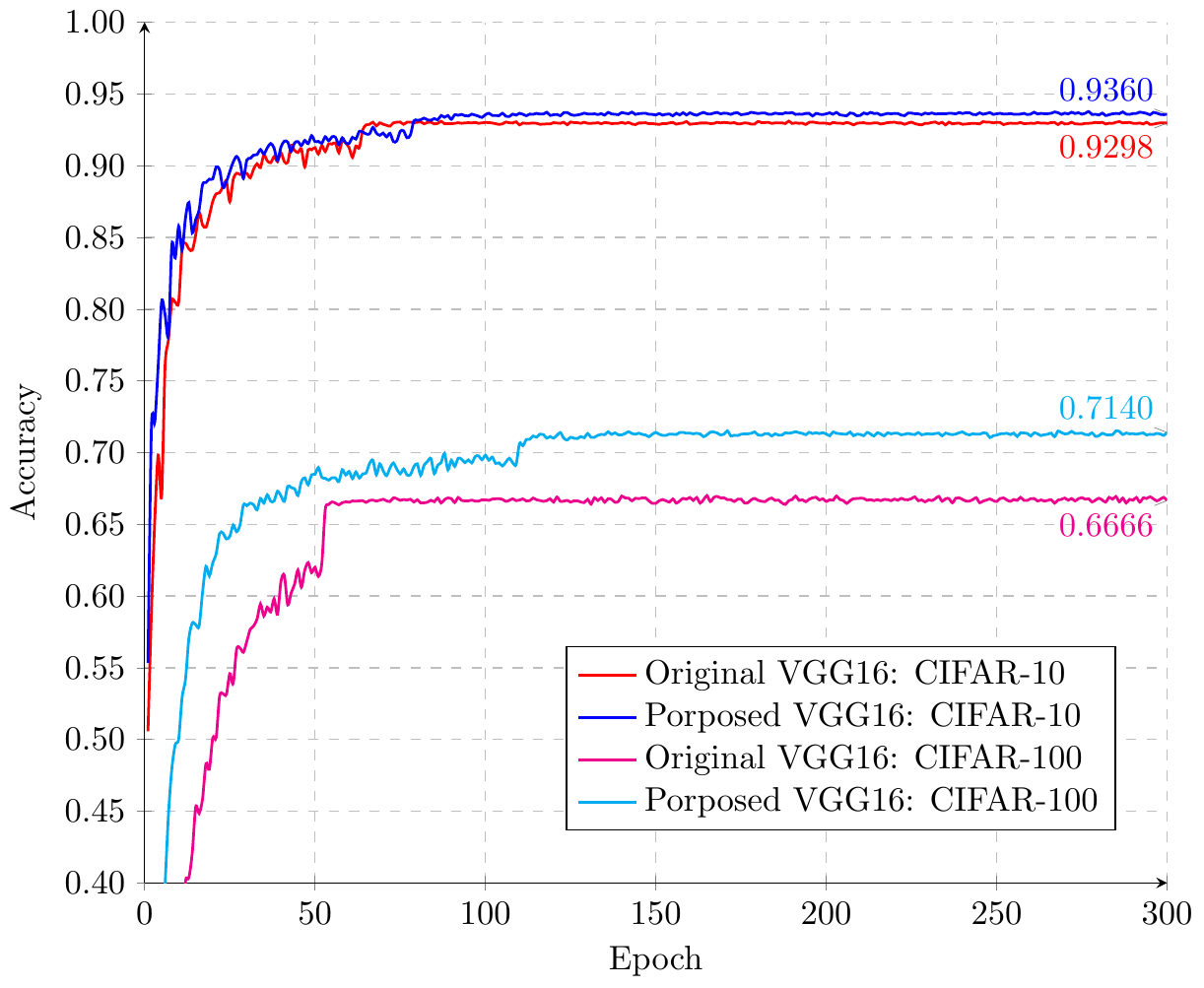}
	\caption{The accuracy of training data after 300 epoch in the original and proposed VGG16.}
	\label{fig:VGG16}
\end{figure}
\begin{figure} [h]
	\centering
	\includegraphics{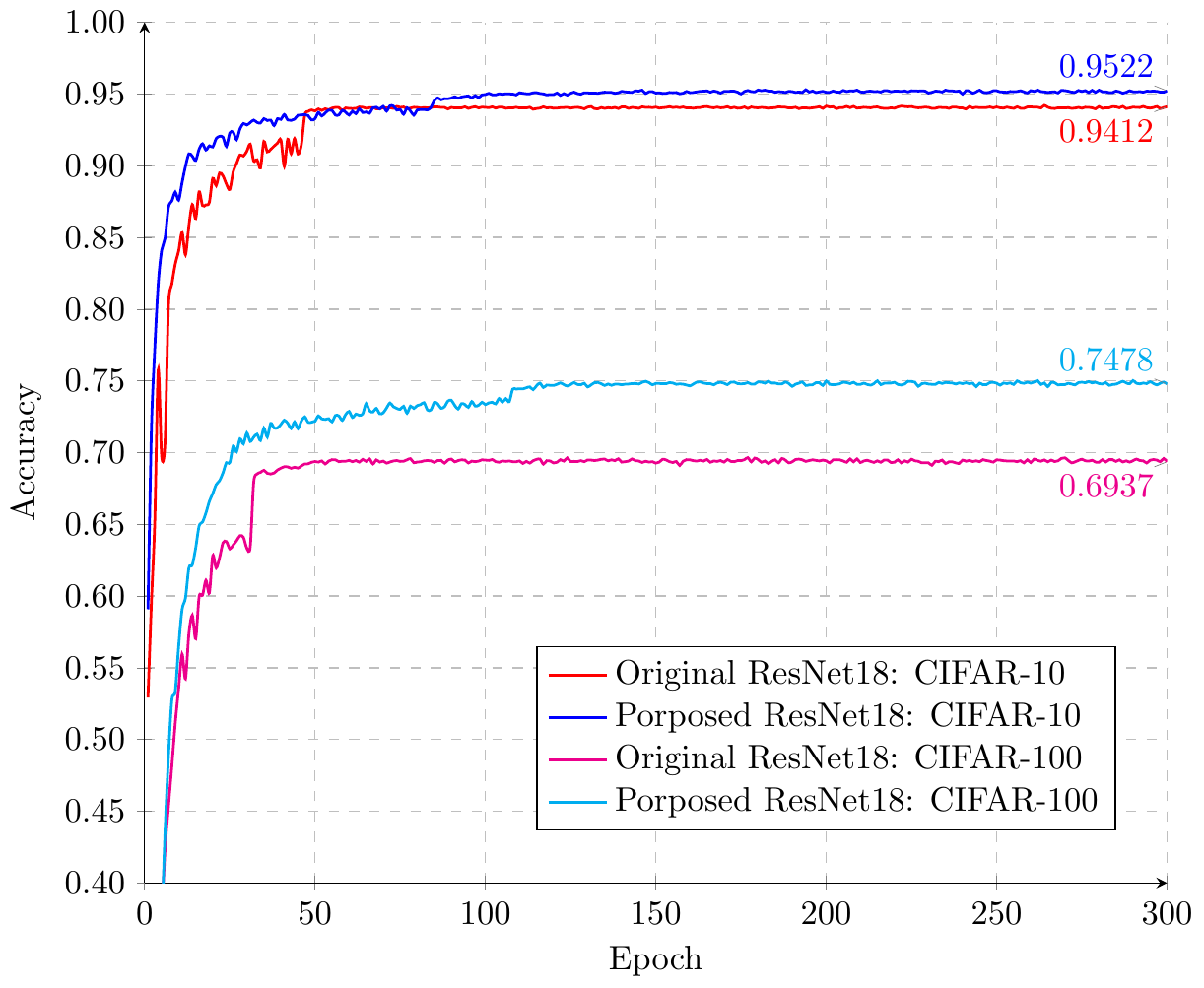}
	\caption{The accuracy of training data after 300 epoch in the original and proposed ResNet18.}
	\label{fig:ResNet18}
\end{figure}
\begin{figure} [h]
\centering
\includegraphics{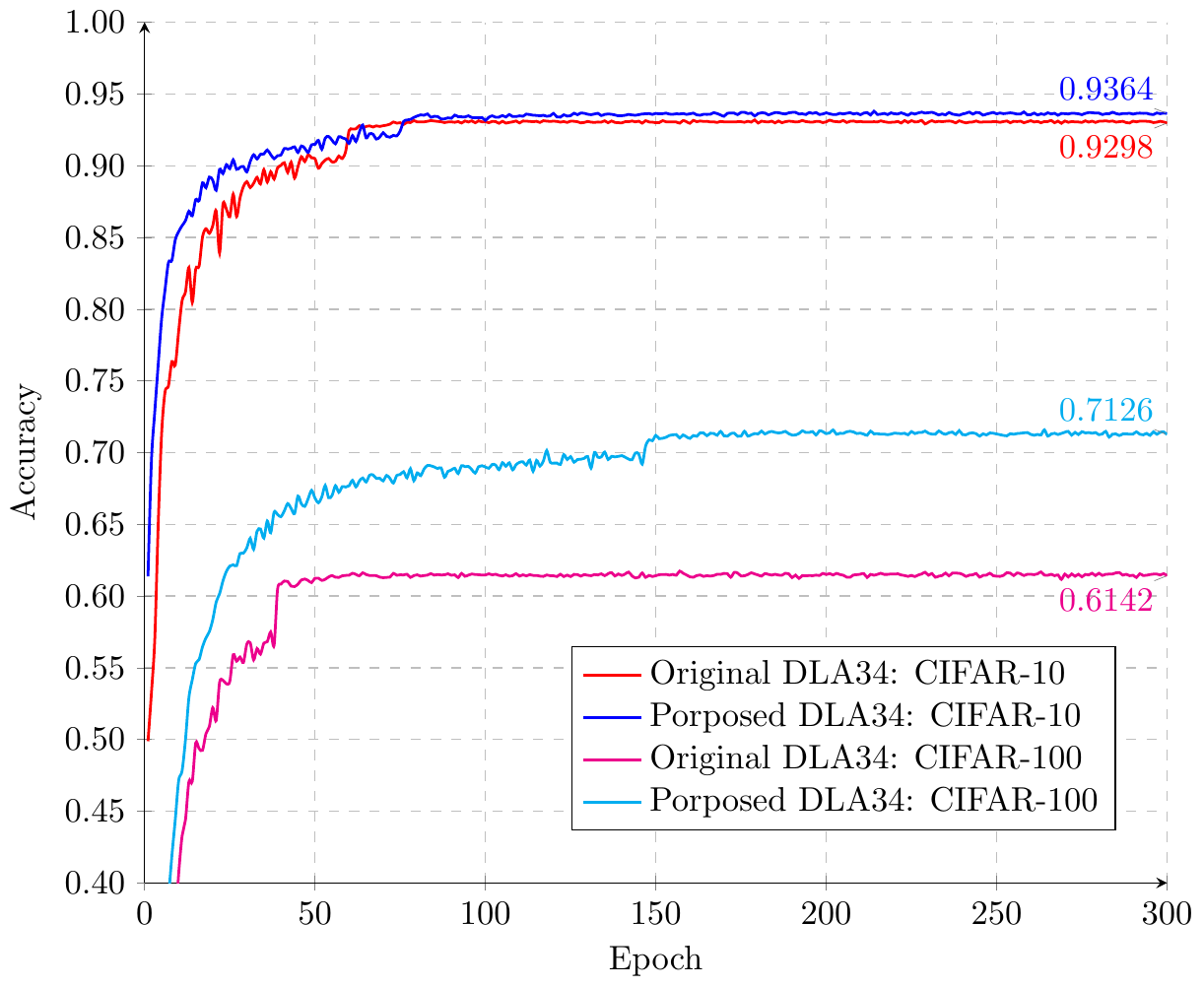}
\caption{The accuracy of training data after 300 epoch in the original and proposed DenseNet181.}
\label{fig:DLA34}
\end{figure}
\begin{figure} [h]
\centering
\includegraphics{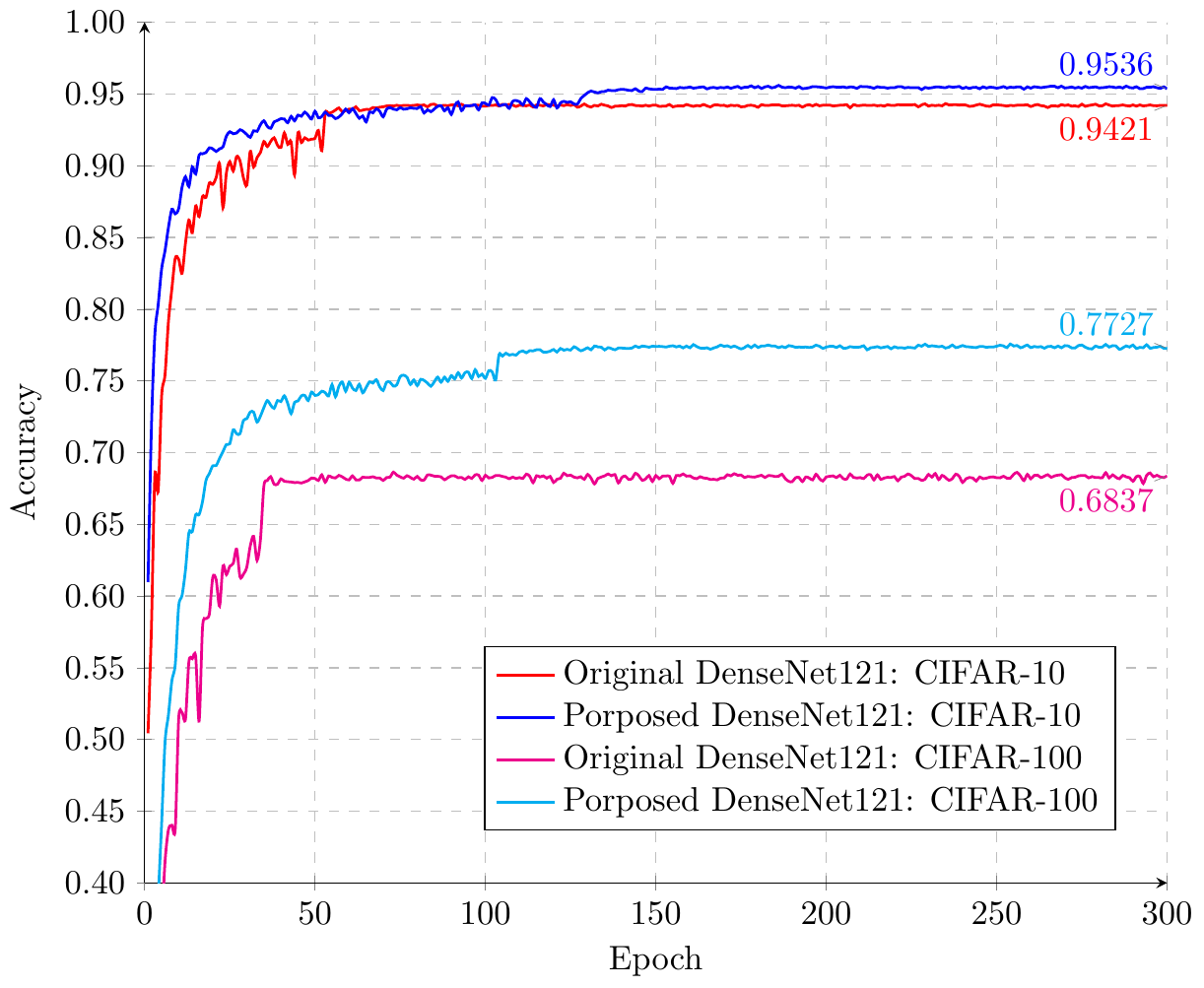}
\caption{The accuracy of training data after 300 epoch in the original and proposed DenseNet181.}
\label{fig:DenseNet181}
\end{figure}

For more credibility, each model has been tested eight times.
However, we only show the results with the best accuracy in the figures from \autoref{fig:VGG16} to \autoref{fig:DenseNet181},
with the purpose to visualize the convergence during the training period.
As expected in \autoref{subsec:ScoreNormalization}, we see that the convergence contributed by the proposed method is significantly improved.
They converge faster with increasing accuracy.
All the experiments can be well trained within 150 epochs.
Note that in addition to the (best) results shown in these figures,
the convergences of all the experiments using the proposed method are better than those of the original when the accuracy is greater than 0.5.
On the other hand, \autoref{tab:original} and \autoref{tab:MC} show that the proposed classifier requires more training parameters,
and the FLOPS also increases, but the increase of the FLOPS can almost overcome the time increase caused by the increased parameters,
so this is only a little more than the original time.

\begin{table} [h]
	\centering
	\caption{The average accuracy of CIFAR-10 test data with the error range.}
	\begin{tabular}{c|c|c|c|c}
		\toprule
		\textbf{\textit{Method}} & \textbf{VGG16} & \textbf{ResNet18} & \textbf{DLA34} & \textbf{DenseNet121} \\
		\midrule
		\textit{Original} & $0.9258\pm0.0031$ & $0.9334\pm0.0042$ & $0.9205\pm0.0032$ & $0.9359\pm0.0037$ \\
		\midrule
		\textit{Proposed} & $0.9336\pm0.0005$ & $0.9505\pm0.0013$ & $0.9337\pm0.0021$ & $0.9528\pm0.0029$ \\
		\bottomrule
	\end{tabular}
	\label{tab:averaygeCIFAR10}
\end{table}
\begin{table} [h]
\centering
\caption{The average accuracy of CIFAR-100 test data with the error range.}
\begin{tabular}{c|c|c|c|c}
	\toprule
	\textbf{\textit{Method}} & \textbf{VGG16} & \textbf{ResNet18} & \textbf{DLA34} & \textbf{DenseNet121} \\
	\midrule
	\textit{Original} & $0.6626\pm0.0033$ & $0.6933\pm0.0063$ & $0.6056\pm0.0124$ & $0.6815\pm0.0069$ \\
	\midrule
	\textit{Proposed} & $0.7174\pm0.0036$ & $0.7450\pm0.0061$ & $0.7185\pm0.0027$ & $0.7669\pm0.0052$ \\
	\bottomrule
\end{tabular}
\label{tab:averaygeCIFAR100}
\end{table}

Meanwhile, \autoref{tab:averaygeCIFAR10} and \autoref{tab:averaygeCIFAR100} list the average accuracy of the test data with the error range,
with the purpose to illustrate the improvement by our method.
It can be seen that the accuracy and the stability in our method have been significantly improved.
The accuracy by the proposed method is found to improve the overall classification performance in multiple runs.
The most obvious improvement is the CIFAR-100 classification, by a gap of 5.0\%.
Commonly, with the discovery of VGG16, ResNet18, DLA34 and DensNet121,
the performance improves, and our method can further improve these functions without a lot of work.
For more complex and deeper networks, the gain of our method will be even higher.
In addition to these network backbones, our experiment also requires more pre-processing and post-processing,
the reason for that is less relevant and we are not going into the detail.
Please find the complete source code and experimental results in the supplemental files.

\section{Conclusion}
\label{sec:Conclusion}

We present a multiple classifier method that can improve the performance from a new perspective,
and as the number of modules multiplying their connectivity get greater, the method is more effective.
By adjusting the concatenated architectures used for classification tasks,
we identify the need for multiple classifiers to participate, and make the final decision according to their results.
We further discover the condition to enhance the convergence, and embed it into the proposed classifier.
Compared with the original model, our method is more accurate, and can make use of parameters and computations more efficiently.
Experiments show that the dominant architectures can all be improved by using the multiple classifiers.
The gap of accuracy improvement is obvious.

\bibliographystyle{amsrefs}
\bibliography{bib/bibliography}

\begin{thebibliography}{10}
\expandafter\ifx\csname url\endcsname\relax
  \def\url#1{\texttt{#1}}\fi
\expandafter\ifx\csname urlprefix\endcsname\relax\def\urlprefix{URL }\fi
\expandafter\ifx\csname href\endcsname\relax
  \def\href#1#2{#2} \def\path#1{#1}\fi

\bibitem{krizhevsky2012imagenet}
A.~Krizhevsky, I.~Sutskever, G.~E. Hinton, Imagenet classification with deep
  convolutional neural networks, in: {NIPS}, 2012, pp. 1106--1114.

\bibitem{lawrence1997face}
S.~Lawrence, C.~L. Giles, A.~C. Tsoi, A.~D. Back, Face recognition: a
  convolutional neural-network approach, {IEEE} Trans. Neural Networks 8~(1)
  (1997) 98--113.

\bibitem{sun2015deepid3}
Y.~Sun, D.~Liang, X.~Wang, X.~Tang, Deepid3: Face recognition with very deep
  neural networks, CoRR abs/1502.00873 (2015).

\bibitem{ranjan2017all}
R.~Ranjan, S.~Sankaranarayanan, C.~D. Castillo, R.~Chellappa, An all-in-one
  convolutional neural network for face analysis, in: {FG}, {IEEE} Computer
  Society, 2017, pp. 17--24.

\bibitem{shen2017deep}
D.~Shen, G.~Wu, H.-I. Suk, Deep learning in medical image analysis, Annual
  Review of Biomedical Engineering 19~(1) (2017) 221--248.

\bibitem{jiang2010medical}
J.~Jiang, P.~R. Trundle, J.~Ren, Medical image analysis with artificial neural
  networks, Comput. Medical Imaging Graph. 34~(8) (2010) 617--631.

\bibitem{nawaz2018classification}
W.~Nawaz, S.~Ahmed, A.~Tahir, H.~A. Khan, Classification of breast cancer
  histology images using {ALEXNET}, in: {ICIAR}, Vol. 10882 of Lecture Notes in
  Computer Science, Springer, 2018, pp. 869--876.

\bibitem{tian2018deeptest}
Y.~Tian, K.~Pei, S.~Jana, B.~Ray, Deeptest: automated testing of
  deep-neural-network-driven autonomous cars, in: {ICSE}, {ACM}, 2018, pp.
  303--314.

\bibitem{kim2015deep}
D.~K. Kim, T.~Chen, Deep neural network for real-time autonomous indoor
  navigation, CoRR abs/1511.04668 (2015).

\bibitem{gong2014re}
S.~Gong, M.~Cristani, C.~C. Loy, T.~M. Hospedales, The re-identification
  challenge, in: Person Re-Identification, Advances in Computer Vision and
  Pattern Recognition, Springer, 2014, pp. 1--20.

\bibitem{xiao2017scene}
L.~Xiao, Q.~Yan, S.~Deng, Scene classification with improved alexnet model, in:
  {ISKE}, {IEEE}, 2017, pp. 1--6.

\bibitem{wang2015training}
L.~Wang, C.~Lee, Z.~Tu, S.~Lazebnik, Training deeper convolutional networks
  with deep supervision, CoRR abs/1505.02496 (2015).

\bibitem{zhang2016understanding}
C.~Zhang, S.~Bengio, M.~Hardt, B.~Recht, O.~Vinyals, Understanding deep
  learning requires rethinking generalization, in: {ICLR}, OpenReview.net,
  2017.

\bibitem{simonyan2014very}
K.~Simonyan, A.~Zisserman, Very deep convolutional networks for large-scale
  image recognition, in: {ICLR}, 2015.

\bibitem{zhong2020random}
Z.~Zhong, L.~Zheng, G.~Kang, S.~Li, Y.~Yang, Random erasing data augmentation,
  in: {AAAI}, {AAAI} Press, 2020, pp. 13001--13008.

\bibitem{shorten2019survey}
C.~Shorten, T.~M. Khoshgoftaar, A survey on image data augmentation for deep
  learning, J. Big Data 6 (2019) 60.

\bibitem{szegedy2015going}
C.~Szegedy, W.~Liu, Y.~Jia, P.~Sermanet, S.~E. Reed, D.~Anguelov, D.~Erhan,
  V.~Vanhoucke, A.~Rabinovich, Going deeper with convolutions, in: {CVPR},
  {IEEE} Computer Society, 2015, pp. 1--9.

\bibitem{he2016deep}
K.~He, X.~Zhang, S.~Ren, J.~Sun, Deep residual learning for image recognition,
  in: {CVPR}, {IEEE} Computer Society, 2016, pp. 770--778.

\bibitem{krizhevsky2009learning}
A.~Krizhevsky, G.~Hinton, et~al., Learning multiple layers of features from
  tiny images (2009).

\bibitem{bridle1990probabilistic}
J.~S. Bridle, Probabilistic interpretation of feedforward classification
  network outputs, with relationships to statistical pattern recognition, in:
  {NATO} Neurocomputing, Vol.~68 of {NATO} {ASI} Series, Springer, 1989, pp.
  227--236.

\bibitem{dietterich1994solving}
T.~G. Dietterich, G.~Bakiri, Solving multiclass learning problems via
  error-correcting output codes, J. Artif. Intell. Res. 2 (1995) 263--286.

\bibitem{zhou2012coding}
J.-D. Zhou, X.-D. Wang, H.-J. Zhou, Y.-H. Cui, S.~Jing, Coding design for error
  correcting output codes based on perceptron, Optical Engineering 51~(5)
  (2012) 1--7.

\bibitem{ciresan2011convolutional}
D.~C. Ciresan, U.~Meier, L.~M. Gambardella, J.~Schmidhuber, Convolutional
  neural network committees for handwritten character classification, in:
  {ICDAR}, {IEEE} Computer Society, 2011, pp. 1135--1139.

\bibitem{vapnik1999overview}
V.~Vapnik, An overview of statistical learning theory, {IEEE} Trans. Neural
  Networks 10~(5) (1999) 988--999.

\bibitem{joachims1998making}
T.~Joachims, Making large-scale svm learning practical, Technical report,
  Dortmund (1998).

\bibitem{byun2003survey}
H.~Byun, S.~Lee, A survey on pattern recognition applications of support vector
  machines, Int. J. Pattern Recognit. Artif. Intell. 17~(3) (2003) 459--486.

\bibitem{naranjo2017two}
L.~Naranjo, C.~J. Perez, J.~Mart{\'{\i}}n, Y.~Campos{-}Roca, A two-stage
  variable selection and classification approach for parkinson's disease
  detection by using voice recording replications, Comput. Methods Programs
  Biomed. 142 (2017) 147--156.

\bibitem{elleuch2016new}
M.~Elleuch, R.~Maalej, M.~Kherallah, A new design based-svm of the {CNN}
  classifier architecture with dropout for offline arabic handwritten
  recognition, in: {ICCS}, Vol.~80 of Procedia Computer Science, Elsevier,
  2016, pp. 1712--1723.

\bibitem{hershey2017cnn}
S.~Hershey, S.~Chaudhuri, D.~P.~W. Ellis, J.~F. Gemmeke, A.~Jansen, R.~C.
  Moore, M.~Plakal, D.~Platt, R.~A. Saurous, B.~Seybold, M.~Slaney, R.~J.
  Weiss, K.~W. Wilson, {CNN} architectures for large-scale audio
  classification, in: {ICASSP}, {IEEE}, 2017, pp. 131--135.

\bibitem{ball2017comprehensive}
J.~E. Ball, D.~T. Anderson, C.~S. Chan, A comprehensive survey of deep learning
  in remote sensing: Theories, tools and challenges for the community, CoRR
  abs/1709.00308 (2017).

\bibitem{mandelbaum2017distance}
A.~Mandelbaum, D.~Weinshall, Distance-based confidence score for neural network
  classifiers, CoRR abs/1709.09844 (2017).

\bibitem{ueda2000optimal}
N.~Ueda, Optimal linear combination of neural networks for improving
  classification performance, {IEEE} Trans. Pattern Anal. Mach. Intell. 22~(2)
  (2000) 207--215.

\bibitem{lai2015learning}
K.~Lai, D.~Liu, S.~Chang, M.~Chen, Learning sample specific weights for late
  fusion, {IEEE} Trans. Image Process. 24~(9) (2015) 2772--2783.

\bibitem{cao2018dynamic}
H.~Cao, S.~Bernard, L.~Heutte, R.~Sabourin, Dynamic voting in multi-view
  learning for radiomics applications, in: {S+SSPR}, Vol. 11004 of Lecture
  Notes in Computer Science, Springer, 2018, pp. 32--41.

\bibitem{diez2015diversity}
J.~D{\'{\i}}ez{-}Pastor, J.~J. Rodr{\'{\i}}guez, C.~I. Garc{\'{\i}}a{-}Osorio,
  L.~I. Kuncheva, Diversity techniques improve the performance of the best
  imbalance learning ensembles, Inf. Sci. 325 (2015) 98--117.

\bibitem{wu2019wider}
Z.~Wu, C.~Shen, A.~van~den Hengel, Wider or deeper: Revisiting the resnet model
  for visual recognition, Pattern Recognit. 90 (2019) 119--133.

\bibitem{hammann2020identity}
M.~Hammann, M.~Kraus, S.~Shafaei, A.~C. Knoll, Identity recognition in
  intelligent cars with behavioral data and lstm-resnet classifier, CoRR
  abs/2003.00770 (2020).

\bibitem{yunpeng2017sharing}
Y.~Chen, X.~Jin, B.~Kang, J.~Feng, S.~Yan, Sharing residual units through
  collective tensor factorization in deep neural networks, CoRR abs/1703.02180
  (2017).

\bibitem{xie2017aggregated}
S.~Xie, R.~B. Girshick, P.~Doll{\'{a}}r, Z.~Tu, K.~He, Aggregated residual
  transformations for deep neural networks, in: {CVPR}, {IEEE} Computer
  Society, 2017, pp. 5987--5995.

\bibitem{huang2017densely}
G.~Huang, Z.~Liu, L.~van~der Maaten, K.~Q. Weinberger, Densely connected
  convolutional networks, in: {CVPR}, {IEEE} Computer Society, 2017, pp.
  2261--2269.

\bibitem{lecun1998gradient}
Y.~LeCun, L.~Bottou, Y.~Bengio, P.~Haffner, Gradient-based learning applied to
  document recognition, Proceedings of the IEEE 86~(11) (1998) 2278--2324.

\bibitem{ioffe2015batch}
S.~Ioffe, C.~Szegedy, Batch normalization: Accelerating deep network training
  by reducing internal covariate shift, in: {ICML}, Vol.~37 of {JMLR} Workshop
  and Conference Proceedings, JMLR.org, 2015, pp. 448--456.

\bibitem{yu2018deep}
F.~Yu, D.~Wang, E.~Shelhamer, T.~Darrell, Deep layer aggregation, in: {CVPR},
  {IEEE} Computer Society, 2018, pp. 2403--2412.

\bibitem{paszke2017automatic}
A.~Paszke, S.~Gross, S.~Chintala, G.~Chanan, E.~Yang, Z.~DeVito, Z.~Lin,
  A.~Desmaison, L.~Antiga, A.~Lerer, Automatic differentiation in pytorch
  (2017).

\bibitem{kingma2014adam}
D.~P. Kingma, J.~Ba, Adam: {A} method for stochastic optimization, in: {ICLR}
  (Poster), 2015.

\end{thebibliography}


\end{document}